\documentclass{article}

\usepackage{nac_preprint}
\usepackage[utf8]{inputenc} 
\usepackage[T1]{fontenc}    
\usepackage{hyperref}       
\usepackage{url}            
\usepackage{booktabs}       
\usepackage{amsfonts}       
\usepackage{nicefrac}       
\usepackage{microtype}      
\usepackage{tabularx}
\usepackage{booktabs} 
\usepackage{subcaption}
\usepackage{changepage}
\usepackage{amsmath}
\usepackage{amssymb}
\usepackage{cleveref}
\usepackage{mathtools}
\usepackage[toc,page]{appendix}
\usepackage{enumitem}
\usepackage{graphics}
\usepackage{graphicx}
\usepackage{xcolor}
\usepackage[export]{adjustbox}
\usepackage{algorithm}
\usepackage{algorithmicx}
\usepackage[noend]{algpseudocode}
\algnewcommand{\LineComment}[1]{\State \(//\) #1}
\algnewcommand{\RLineComment}[1]{\State \(\triangleright\) #1}
\usepackage{bm}
\usepackage{multirow}

\usepackage{wrapfig}

\newcolumntype{C}[1]{>{\hsize=#1\hsize\centering\arraybackslash}X}

\usepackage{etoolbox}
\usepackage{tikz}
\usetikzlibrary{tikzmark}
\usetikzlibrary{calc}

\errorcontextlines\maxdimen

\newcommand{\ALGtikzmarkcolor}{black}
\newcommand{\ALGtikzmarkextraindent}{4pt}
\newcommand{\ALGtikzmarkverticaloffsetstart}{-.5ex}
\newcommand{\ALGtikzmarkverticaloffsetend}{-.5ex}
\makeatletter
\newcounter{ALG@tikzmark@tempcnta}

\newcommand\ALG@tikzmark@start{%
    \global\let\ALG@tikzmark@last\ALG@tikzmark@starttext%
    \expandafter\edef\csname ALG@tikzmark@\theALG@nested\endcsname{\theALG@tikzmark@tempcnta}%
    \tikzmark{ALG@tikzmark@start@\csname ALG@tikzmark@\theALG@nested\endcsname}%
    \addtocounter{ALG@tikzmark@tempcnta}{1}%
}

\def\ALG@tikzmark@starttext{start}
\newcommand\ALG@tikzmark@end{%
    \ifx\ALG@tikzmark@last\ALG@tikzmark@starttext
    \else
        \tikzmark{ALG@tikzmark@end@\csname ALG@tikzmark@\theALG@nested\endcsname}%
        \tikz[overlay,remember picture] \draw[\ALGtikzmarkcolor] let \p{S}=($(pic cs:ALG@tikzmark@start@\csname ALG@tikzmark@\theALG@nested\endcsname)+(\ALGtikzmarkextraindent,\ALGtikzmarkverticaloffsetstart)$), \p{E}=($(pic cs:ALG@tikzmark@end@\csname ALG@tikzmark@\theALG@nested\endcsname)+(\ALGtikzmarkextraindent,\ALGtikzmarkverticaloffsetend)$) in (\x{S},\y{S})--(\x{S},\y{E});%
    \fi
    \gdef\ALG@tikzmark@last{end}%
}

\apptocmd{\ALG@beginblock}{\ALG@tikzmark@start}{}{\errmessage{failed to patch}}
\pretocmd{\ALG@endblock}{\ALG@tikzmark@end}{}{\errmessage{failed to patch}}
\makeatother

\title{Uncertainty-Driven Replay Memory for Reinforcement Learning}

\author{%
Sheeraja Rajakrishnan \\
Rochester Institute of Technology \\
\texttt{sr8685@rit.edu}
\And
Alexander G. Ororbia \\
Rochester Institute of Technology \\
\texttt{ago@cs.rit.edu}
\AND 
Travis Desell \\
Rochester Institute of Technology \\ 
\texttt{tjdvse@rit.edu}
\And
Daniel E. Krutz \\
Rochester Institute of Technology \\ 
\texttt{dxkvse@rit.edu}
}

\begin{document}

\setlength{\abovedisplayskip}{0.065cm}
\setlength{\belowdisplayskip}{0pt}

\maketitle

\begin{abstract}{ 
Uncertainty estimation provides promising capabilities for reinforcement learning (RL) agents. Notably, estimating uncertainty can reduce the training time and enable agents to obtain greater rewards over time by exploiting information related to whether an action would facilitate exploration of portions of an environment that are well-known versus those that are relatively unknown. In this work, we propose a novel formulation of the experience replay buffer commonly used in RL that we call \emph{uncertainty-driven replay memory} (UDRM), which entails an update scheme for internally stored memories based on uncertainty estimates obtained by an RL model during training. In contrast to existing forms of RL, which typically use temporal difference error or the distribution of transitions to update the replay memory buffer and train RL controllers, our scheme biases the memory buffer to store more uncertain transitions that will improve an RL agent's generalization throughout training. Experimental results demonstrate that our proposed uncertainty-aware replay buffer enables an RL agent to obtain higher rewards during training compared to other existing uncertainty-aware RL frameworks.
}  

\keywords{Reinforcement Learning \and Uncertainty Estimation \and Experience Replay \and Memory}
\end{abstract}

\section{Introduction}
\label{sec:intro}

Predictions made by statistical learning models can be unreliable and are frequently overconfident~\cite{lakshminarayanan2017simple-4dc,malinin2018predictive-f84,zhang2021dense-4ed}. This is an issue in reinforcement learning (RL) and one that uncertainty-aware models are equipped to tackle, leading to improved performance and trustworthiness~\cite{silva2020uncertainty-aware-cf4,zhang2020robust-a96}. In RL, agents often make use of a memory replay buffer to learn from prior experience(s) \cite{mnih2013playing-5af}. 
However, many existing frameworks populate this replay buffer based on recency (for example, implementing a ring buffer data structure for storing transitions), temporal difference (TD) error \cite{schaul2015prioritized-85b}, or with respect to certain features of the transitions/states. In this work, we propose a modified approach for populating the replay buffer based on the uncertainty as estimated by an agent model's action selection mechanism. 
In essence, in our framework, using uncertainty enables the RL model to make more optimal decisions when there is limited information about the model's action space. In this work, we make the following key contributions:
\begin{itemize}
    \item We introduce a novel uncertainty-driven RL framework 
    where an agent's memory buffer is populated based on its prediction uncertainty.
    \item We provide empirical evidence that our proposed memory model, UDRM, obtains higher scores during training and higher overall average cumulative return when compared with UADQN~\cite{clements2019estimating-07f} and CEQR-DQN~\cite{stutts2024echoes-08b} baselines.
    \item We demonstrate the performance of UDRM on MinAtar~\cite{young2019minatar-eaa}, Classic Control, and Toy Text Gym environments~\cite{towers2024gymnasium-55d}, to show the applicability of our model across varied environments.
\end{itemize}

\section{Related Work}
\label{sec:related_work}

Machine learning (ML) models can confidently provide incorrect predictions for situations involving data that is not encountered during training~\cite{zhang2021dense-4ed}. When working with partial or incomplete information, the process of determining how likely a particular outcome(s) might be involves computing two key forms of uncertainty: aleatoric and epistemic uncertainty. 
Aleatoric (stochastic) uncertainty quantifies the unknowns that differ each time that a specific experiment is repeated; it is inherent to the input data and cannot be reduced with training an ML model further/longer. 
Epistemic (systematic) uncertainty stems from insufficient knowledge (e.g., inaccurate measurements, not capturing certain effects within a model, etc.) and can, in contrast, 
be reduced by training an ML model longer and with varied training data. 
Existing research uses different methods to estimate uncertainty; for instance, ensemble learning~\cite{liu2019accurate-a71} and entropy-variance estimation schemes~\cite{depeweg2018decomposition-070}. These methods require multiple artificial neural networks (ANNs) or multiple passes through an ANN, often resulting in additional computational and resource overheads.

\subsection{Evidential Deep Learning} 
\label{sec:edl}

Evidential deep learning (EDL) models provide aleatoric and epistemic uncertainty estimates in one pass and do not require the overhead of multiple ANNs, unlike ensemble systems~\cite{amini2020deep-c4c,jrgens2024is-13f,malinin2018predictive-f84,meinert2022unreasonable-a04,sensoy2018evidential-1d4}. In essence, EDL places an evidential prior distribution over the (model) likelihood function, trains the ANN, and obtains the hyper-parameters of the resulting evidential distribution. The evidence collected for a prediction is then calculated based on the Dempster-Shafer theory of evidence~\cite{sentz2002combination-059}. In classification, each possible state output is assigned a ``mass'' (with the total mass summing/integrating to one). These mass values are assigned based on the evidence available to support a particular state, usually obtained from the output of the ANN. The outcome is considered to be ambiguous if all of the states are given an equal mass. An ML model then assigns higher uncertainty values to such predictions since it is not able to decide on one particular output. Alternatively, if the mass is high for one of the states, then the evidence is higher for that state and the model should assign a lower uncertainty for its prediction. In the case of regression, aleatoric and epistemic uncertainty values are calculated based on the inferred hyper-parameters of the evidential distribution. Existing literature has focused on improving the performance of these evidential models~\cite{httel2023deep-a01,meinert2022unreasonable-a04,wu2024evidence-f4b}. We follow Stutts et al.~\cite{stutts2024echoes-08b} and adopt the EDL framing for estimating uncertainty.

\subsection{Uncertainty in Reinforcement Learning} 
\label{sec:uncertainty_rl}

Uncertainty estimation has also proven useful in improving the performance of RL agents~\cite{charpentier2022disentangling-a34,ibrahim2023uncertainty-54f,lockwood2022review-c05,nguyen2022how-dd0,prez2023discrete-ba1,zhao2019uncertainty-based-5ea}. UADQN, proposed by~\cite{clements2019estimating-07f}, employs quantile regression to estimate aleatoric as well as epistemic uncertainty. CEQR-DQN~\cite{stutts2024echoes-08b} is an improvement over the UADQN model, making use of EDL to estimate the uncertainty. This estimated uncertainty is then used to guide exploration. The UBER~\cite{remonda2025uncertainty-based-f92} framework introduces an uncertainty-based filtering scheme before populating the replay memory buffer. In addition, the Wasserstein distance (function) has been used as a measure of uncertainty~\cite{bellemare2017distributional-616} -- if the uncertainty is greater than a set threshold, the transition is stored within the replay buffer. This ensures the replay buffer stores useful memories based on the uncertainty estimates. 

In contrast to UBER, our model does not impede the addition of any transitions but instead injects additional transitions (entries) that the RL model is uncertain about. This core design mechanism allows the model to sample from transitions that are more uncertain while further permitting the sampling of certain transitions; this promotes exploration. 
By the end of the overall training loop, as the RL model becomes more certain, the uncertainty eventually decreases and the replay buffer stops receiving duplicate transitions (as it becomes populated by the transitions that the RL controller is most certain about); this promotes exploitation towards the end of training. 
UPER~\cite{carrasco-davis2025uncertainty-908} proposes a replay buffer prioritization scheme based on uncertainty -- their scheme's estimated uncertainty is used to compute an information gain criterion for prioritizing transitions within the buffer. However, our proposed UDRM offers a simpler, faster mechanism for adding and sampling transitions from the replay memory buffer, building on and generalizing the CEQR-DQN base scheme.

\subsection{Replay Memory Buffers}
\label{sec:replay_memory}

Memory buffers, as proposed by Lin et al.~\cite{lin1992self-improving-ee2}, allow RL agents to replay experiences. Current research has been focused on fine-tuning the replay buffer as well as designing techniques to sample from them in order to train an RL agent. Usefully, Fedus et al.~\cite{fedus2020revisiting-e16} demonstrate the effect of increasing replay buffer capacity on performance; nevertheless, existing research has also demonstrated that large replay memory buffers can negatively impact an RL agent's performance~\cite{eren2023importance-5c1,zhang2017deeper-550}. When storing transition memories in a buffer, Li et al.~\cite{li2021revisiting-dd9} demonstrate that the relationship between the temporal difference (TD) error and the importance of a transition is critical to consider. Bruin et al.~\cite{bruin2018experience-3b9} use characteristics such as age and TD error to determine which transitions that are to be retained in the memory buffer. In contrast, Oh et al.~\cite{oh2020learning-7e8} design a permutation-equivariant neural system -- the neural experience replay sampler (NERS) -- which uses features from transitions to learn contexts; this ``context awareness'' allows NERS to sample meaningful transitions. ``Remember and Forget for Experience Replay''~\cite{novati2019remember-b8a} distinguishes incoming transitions-to-store as either near-policy or far-policy based on their probability of being selected by the current behavior(al) policy. In the interest of pruning the memory buffer carefully, Rahimi et al.~\cite{rahimi-kalahroudi2023replay-f98} propose mechanisms for removing local neighborhood samples from the replay buffer whereas Eysenbach et al.~\cite{eysenbach2019search-2af} employ a graph search over the replay buffer to identify subgoals (this was found to improve agent performance in sparse reward settings).

Prioritized experience replay (PER)~\cite{schaul2015prioritized-85b} centrally uses TD error to prioritize its actions and performs prioritization based on the proportions and ranks of the transitions to be stored within the replay buffer. However, PER has been shown to lead to worse performances in OpenAI Gym environments~\cite{pan2020understanding-5d1,sujit2022prioritizing-c88}; this reduced performance has been attributed to instability in training, outdated priorities, and insufficient sample space coverage. PER samples from unlearnable transitions if the TD error is high, thereby rendering the RL model unable to solve the task. Fujimoto et al.~\cite{fujimoto2020equivalence-87e} propose updated/alternative loss functions for PER whereas Horgan et al.~\cite{horgan2018distributed-dcc} adopt PER in a distributed learning environment. Zha et al.~\cite{zha2019experience-4e8} propose experience replay optimization which updates the agent and replay policies through a return-driven mechanism -- the replay policy is updated based on the cumulative rewards and the subsets are sampled based on a priority vector. Model-augmented PER (MaPER), proposed by Oh et al.~\cite{oh2022model-augmented-f7f}, directly modifies the critic network to predict the next state and reward in addition to the Q-value itself. Attentive experience replay (AER)~\cite{sun2020attentive-163} prioritizes transitions based on the frequency of visits to its state according to the current policy. 

In contrast to the above, hindsight experience replay (HER)~\cite{andrychowicz2017hindsight-913} and competitive experience replay (CER)~\cite{liu2019competitive-068} focus on sparse-reward environments. While HER handles learning by labeling visited states as goals, the latter scheme -- CER --  uses two agents that compete with one another to improve overall system learning. Dynamic experience replay~\cite{luo2020dynamic-446} utilizes multiple replay buffers from which the agent can sample whereas Brittain et al.~\cite{brittain2019prioritized-53d} propose a scheme that prioritizes sequences of experience(s) in order to efficiently learn and improve generalization ability. Our proposed UDRM differs from these existing replay memory buffer variants in that it uses epistemic uncertainty to directly sample transitions that are more effective in training the RL agent; it does not have any additional overhead of calculating weights or other metrics for prioritizing training samples. UDRM is simple and efficient, and our experiments demonstrate its efficacy.

\section{Methodology}
\label{sec:methods}

\begin{figure}[!th]
    \centering
    \includegraphics[width=0.9\linewidth]{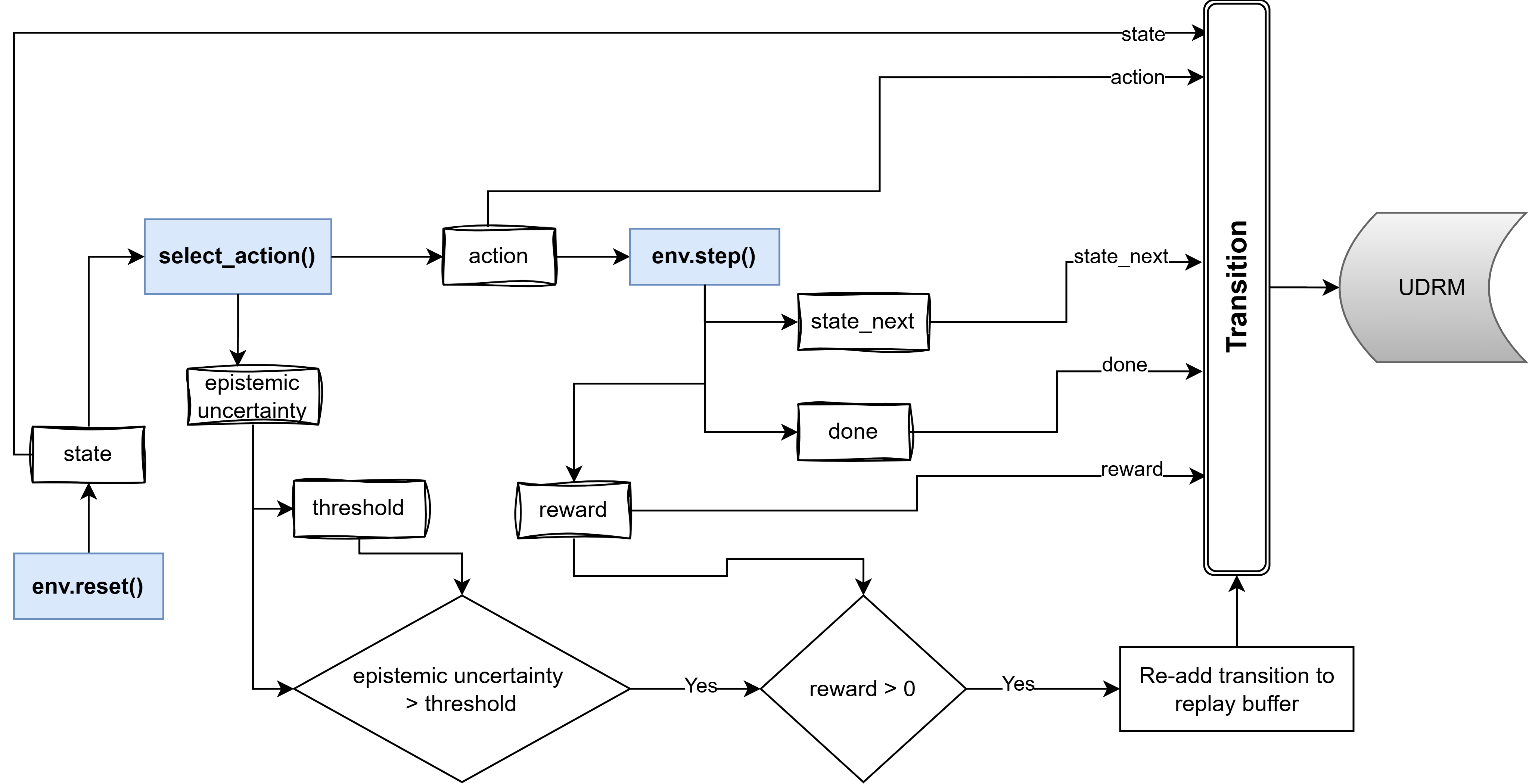}
    \caption{Architecture of the proposed uncertainty-driven replay memory (UDRM) model.}
    \label{fig:UDRMArch_Explore}
\end{figure}

While existing research proposes updates to replay memory based on TD error or a transition's position within the buffer~\cite{liu2019competitive-068,schaul2015prioritized-85b,sun2020attentive-163}, our method -- see Figure \ref{fig:UDRMArch_Explore} -- focuses on utilizing uncertainty to update the replay memory buffer. Updating replay memory based on epistemic uncertainty can greatly aid an RL agent to prioritize uncertain states, which improves exploration and learning efficiency. Our proposed scheme starts by guiding the agent towards greater exploration and gradually guides it towards more exploitative behavior by the end of training. This is done by starting the training process by extracting more uncertain training samples and ending by focusing on more certain sample transitions (from memory).

\subsection{Uncertainty Estimation}
\label{sec:uncertainty_estimation}

Again, as mentioned before, EDL considers the ANN training process as an evidence-gathering process~\cite{amini2020deep-c4c,malinin2018predictive-f84,sensoy2018evidential-1d4}, placing a higher-order evidential distribution over the likelihood function in order to allow direct estimation of global uncertainty. The more evidence that is obtained by the ANN, the lower its uncertainty will be. Following Stutts et al.~\cite{stutts2024echoes-08b}, we adopt deep evidential regression~\cite{amini2020deep-c4c} to estimate uncertainty; a normal-inverse-gamma (NIG) distribution is fit over the probabilities of a deep Q-network (DQN) that is equipped with a probability density function. With trainable parameters representing the expected mean ($\gamma$), the precision ($\nu$), the shape ($\alpha$), and the scale ($\beta$) of the distribution ~\cite{amini2020deep-c4c,stutts2024echoes-08b} itself, our model computes the following:
\begin{equation}
    p(\mu,\sigma^2|\gamma,\nu,\alpha,\beta) = \frac{\beta^\alpha\sqrt{\nu}}{\Gamma(\alpha)\sqrt{2\pi\sigma^2}} \Bigg(\frac{1}{\sigma^2}\Bigg)^{\alpha+1} exp\Bigg\{-\frac{2\beta + \nu(\gamma - \mu)^2}{2\sigma^2}\Bigg\}. 
\end{equation}
Based on Amini et al.~\cite{amini2020deep-c4c} and H{\"u}ttel et al.~\cite{httel2023deep-a01}, total evidence, $\ \phi=2\nu+\alpha+\frac{1}{\beta}$, is minimized via a regularized loss function that penalizes high-confidence predictions and increases (model) uncertainty for incorrect predictions. 
The regularization term notably scales evidence to the absolute error; for quantile regression, the titled (pinball) loss is used to scale evidence as follows: 
\begin{equation}
    \mathcal{L}_{reg} = \phi[\mathbb{I}(y \geq \hat{y})q(y - \hat{y}) + \mathbb{I}(y < \hat{y})(1 - q)(\hat{y} - y)]
\end{equation}
where $q$ is predicted target quantile, $y$ is the target, and $\hat{y}$ is the prediction.

The DQN model is then trained by minimizing the following negative log likelihood (NLL)~\cite{amini2020deep-c4c,stutts2024echoes-08b}:
\begin{equation}
    \mathcal{L}_{NLL} = \frac{1}{2}log\Bigg(\frac{\pi}{\nu}\Bigg) - \alpha log(\Omega) + \Bigg(\alpha + \frac{1}{2}\Bigg)log((y - \gamma)^2\nu + \Omega) + log\Bigg(\frac{\Gamma(\alpha)}{\Gamma(\alpha + \frac{1}{2})}\Bigg)
\end{equation}
where$\ \Omega = 2\beta(1 + \nu)$. This NLL loss encourages the DQN to provide NIG distribution parameters that maximize the prediction evidence. The prediction, aleatoric and epistemic uncertainty, are then finally computed in the following manner:
\begin{equation}
    \underbrace{\mathbb{E[\mu]} = \gamma\vphantom{\frac{1}{1}}}_\text{prediction}, \quad\quad \underbrace{\mathbb{E}[\sigma^2] = \frac{\beta}{\alpha - 1}}_\text{aleatoric}, \quad\quad \underbrace{Var[\mu] = \frac{\beta}{\nu(\alpha - 1)}}_\text{epistemic} . 
\end{equation}

\subsection{Uncertainty-Driven Replay Memory (UDRM)}
\label{sec:udrm_buffer}

The core DQN structure we employ is adapted from~\cite{mnih2015human-level-a85} since it has proven to be 
an effective architecture often employed by many baseline models~\cite{clements2019estimating-07f,stutts2024echoes-08b}; building on this structure further facilitates a fair comparison. 
Since we are trying to modify the replay memory buffer, we have designed our uncertainty estimation framework to match that of the CEQR-DQN~\cite{stutts2024echoes-08b} architecture. UA-DQN~\cite{clements2019estimating-07f} employs quantile regression and posterior networks to estimate its uncertainty ove the RL agent's predictions. Since EDL provides uncertainty estimates in a single pass~\cite{amini2020deep-c4c,stutts2024echoes-08b}, we adopt this framework for estimating the epistemic uncertainties of our RL controller's predictions. Our scheme's estimated uncertainty is then used to add transitions to the replay memory.

During the exploration stage, for each transition, the epistemic uncertainty is stored to later calculate the initial alpha and beta values, which are used to exponentially decay an uncertainty threshold value in the UDRM. The threshold for uncertainty, $\xi$, is calculated as a function of the time step $t$: 
\begin{equation}
    \xi = \alpha e^{-\beta t}
\end{equation}
where $\alpha$, the initial value, and $\beta$, the decay constant, are recalibrated at regular intervals. This threshold becomes stricter in later time steps. The initial $\alpha$ is set as the $99$th percentile of uncertainties stored during the pre-training stage; we use $99$th percentile to permit a wider threshold for the uncertainty. This further ensures that the threshold does not zero out too quickly. The initial beta value is calculated as follows: 
\begin{equation}
    \beta_{initial} = \frac{log(k)}{N}
\end{equation}
where $k = 2$ denotes the threshold should be halved by the end of training and $N$ the total number of steps for training the RL agent. During the initial data collection, each transition is directly added to the replay memory. After this data collection stage, some transitions may be added to the replay memory buffer twice based on the uncertainty value of the transition.

\begin{table*}[t]
    \caption{Average Cumulative Reward of the UDRM (Ours) model for the five MinAtar games over 10 trials, compared against UADQN, and CEQR-DQN models. Best results are \textbf{\underline{underlined and in bold}}. Second best results are in \textbf{bold}. UDRM variants outperform or are competitive against other models for MinAtar games.}
    \begin{center}
        \scriptsize
        \setlength{\tabcolsep}{2pt}
        \renewcommand{\arraystretch}{1.5}
        \begin{tabularx}{\textwidth}{|C{0.7}|C{1.0}|C{1.0}|C{1.2}|C{1.1}|}
            \hline
            \textbf{Game} & \textbf{UADQN} & \textbf{CEQR-DQN} & \textbf{UDRM ($\alpha=50$)} & \textbf{UDRM ($\alpha=500$)} \\
            \hline      
            Asterix & 15.0796 $\pm$ 1.4093 & \textbf{34.1909 $\pm$ 3.7789} & 33.7495 $\pm$ 6.4808 & \textbf{\underline{35.1673 $\pm$ 6.0909}} \\
            Breakout & 15.2167 $\pm$ 2.2581 & 13.4228 $\pm$ 4.0635 & \textbf{\underline{22.6026 $\pm$ 12.164}} & \textbf{19.9538 $\pm$ 5.6874} \\
            Freeway & \textbf{\underline{56.7454 $\pm$ 1.6765}} & \textbf{51.3018 $\pm$ 1.8275} & 51.107 $\pm$ 1.1157 & 50.3514 ± 0.9193 \\
            Seaquest & 8.0099 $\pm$ 4.6142 & 19.9233 $\pm$ 4.6946 & \textbf{\underline{28.2662 $\pm$ 3.1341}} & \textbf{22.3905 $\pm$ 4.2818} \\
            SpaceInvaders & 78.4135 $\pm$ 4.4921 & 222.3827 $\pm$ 9.7841 & \textbf{237.5681 $\pm$ 9.7123} & \textbf{\underline{257.8015 $\pm$ 15.5146}} \\
            \hline
        \end{tabularx}
        \label{tab:avg_cum_reward}
    \end{center}
\end{table*}

\begin{table*}[t]
    \caption{Average Cumulative Reward of the UDRM (Ours) model for classic control and toy text games over 10 trials, compared against UADQN, CEQR-DQN, and CEQR-DQN-PER models. Best results are \textbf{\underline{underlined and in bold}}. Second best results are in \textbf{bold}. UDRM variants are competitive against other models for classic control and toy text environments.}
    \begin{center}
        \scriptsize
        \setlength{\tabcolsep}{2pt}
        \renewcommand{\arraystretch}{1.5}
        \begin{tabularx}{\textwidth}{|C{0.7}|C{1.0}|C{1.0}|C{1.0}|C{1.2}|C{1.1}|}
            \hline
           \textbf{Game} & \textbf{UADQN} & \textbf{CEQR-DQN} & \textbf{CEQR-DQN (PER)} & \textbf{UDRM ($\alpha=50$)} & \textbf{UDRM ($\alpha=500$)} \\
            \hline  
            Acrobat & \textbf{\underline{-93.5178 $\pm$ 65.7548}} & \textbf{-110.6482 $\pm$ 83.1804} & -493.6452 $\pm$ 27.6193 & -141.1313 $\pm$ 109.3169 & -130.8336 $\pm$ 92.2315 \\ 
            CartPole & 240.1597 $\pm$ 165.6334 & 260.6963 $\pm$ 100.6745 & 18.5338 $\pm$ 7.2049 & \textbf{290.9214 $\pm$ 132.7345} & \textbf{\underline{294.3896 $\pm$ 138.3297}} \\
            FrozenLake & \textbf{0.4065 $\pm$ 0.4687} & 0.3647 $\pm$ 0.3472 & 0.013 $\pm$ 0.0301 & 0.1985 $\pm$ 0.23 & \textbf{\underline{0.4113 $\pm$ 0.3577}} \\
            MountainCar & \textbf{\underline{-143.1808 $\pm$ 34.0634}} & \textbf{-172.0413 $\pm$ 28.1154} & -200.0 $\pm$ 0.0 & -188.2361 $\pm$ 26.6773 & -190.2514 $\pm$ 21.6079 \\
            \hline
        \end{tabularx}
        \label{tab:classic_toy_reward}
    \end{center}
\end{table*}

\subsubsection{Alpha and Beta Recalibration} 
$\alpha$ and $\beta$, used to drive the exponential decay of the UDRM threshold, are recalibrated at fixed intervals and require prior uncertainty values to be stored. We experiment with two different recalibration intervals for $\alpha$: 
\textit{(1)} every $50$ or $500$ time steps; and, 
\textit{(2)} keeping the $\beta$ recalibration interval at every $5000$ time steps. 
We did not see any drastic change in the performance due to the $\beta$ recalibration interval; hence we provide results only for one interval. The $\alpha$ and $\beta$ recalibration intervals are manually selected; how the recalibration intervals affect different (RL) environments is a direction to be further studied as tuning these hyper-parameters independently for different environments/tasks may result in higher rewards. 

We set a window of $100$ to calculate the moving average of the uncertainty based on which $\alpha$ and $\beta$ are  recalibrated. If the moving average is greater than the threshold, then $\beta$ is decreased by $5$\% to slow the decay process and promote exploration. Similarly, if the moving average is less than the threshold, then $\beta$ is increased by $5$\% to speed up the decay process and promote exploitation. $\alpha$ is recalibrated as the $99$th percentile of the stored epistemic uncertainty values. Future work will examine how aleatoric uncertainty can also be incorporated into this pipeline to further benefit the agent.

\subsubsection{Threshold comparison} 
Since epistemic uncertainty can be reduced by training an ML model further, we use the epistemic uncertainty to compare against the threshold. If the current epistemic uncertainty is greater than the threshold and the reward is greater than zero, then the transition is added to the UDRM buffer; this, again, promotes exploration by skewing the distribution towards uncertain transitions in the replay memory. As the RL controller trains, the epistemic uncertainty will decrease and, as a result, the threshold will also decrease. Towards the end of training, the number of transitions that are re-added to the replay memory buffer will decrease and the buffer will contain more instances of certain transitions. Ultimately, this will lead the agent to explore its environment more at the beginning of training and, gradually, lead the agent towards exploitation towards the end of training. 

We re-inject transitions with a reward of at least one to ensure that only profitable transitions are present twice within the replay memory buffer. When actions are sampled from the modified buffer, the chances of sampling uncertain actions are higher. Note that we used epistemic uncertainty at training time~\cite{charpentier2022disentangling-a34} to sample actions. We demonstrate that this modified approach allows a UDRM-driven RL agent to obtain higher rewards in the same environment within the same number of time steps as compared to other models. Algorithm 1 outlines our UDRM framework, as described above, and Figure~\ref{fig:UDRMArch_Explore} visualizes the architecture of our proposed UDRM system.

\begin{figure*}[htbp]
    \hrule \vskip 3pt
    \noindent \textbf{Algorithm 1:} Proposed Uncertainty-Driven Replay Memory
    \vskip 3pt \hrule \vskip 5pt
    
    \begin{flushleft}        
    
        \textbf{Input:} Replay buffer $\mathcal{D}$, Transition $\tau(s, a, r, s', \text{done})$, Total timesteps $T$, Current timestep $t$, Current transition's epistemic uncertainty $u_{epistemic}$, Initial data collection steps $N_{\mathrm{init}}$, Moving average window $N_{\text{window}}$, Alpha recalibration interval $I_{\alpha}$, Beta recalibration interval $I_{\beta}$ \\
        \textbf{Output:} Updated replay buffer $\mathcal{D}$ \\
        
        \textbf{Initialization:} initialize a list to store historical epistemic uncertainty values $u_{history}$ \\
        \textbf{Initialization:} initialize $\alpha$ = 1, $\beta$ = 1, threshold $\xi$ = 1.0, k = 2, $\beta$ decrease factor $\beta_{dec} = 0.95$, $\beta$ increase factor $\beta_{inc} = 1.05$ \\
        \textbf{for} timestep $t \leftarrow 0$ to $T$ \textbf{do} \\
            Calculate the epistemic uncertainty, $u_{epistemic}$, associated with the action selection for the transition \\
            Execute the selected action, $a$, in the environment and obtain the transition $\tau(s, a, r, s', \text{done})$ \\
            Store the transition $\tau$ in the replay buffer $\mathcal{D}$ \\
            \textbf{if} $t$ $<$ $N_{\mathrm{init}}$ \textbf{then} \\
                \hspace*{1.5em} Append current transition's $u_{epistemic}$ to $u_{history}$ \\
            \textbf{else} \\
                \hspace*{1.5em} \textbf{if} $t$ $==$ $N_{\mathrm{init}}$ \textbf{then} \\
                        \hspace*{3.0em} Calculate: $\alpha \leftarrow Q_{0.99}(u_{history})$ and $beta \leftarrow \frac{log(k)}{T}$ \\                    
                    \hspace*{1.5em} \textbf{else} \\
                        \hspace*{3.0em} Calculate: $\xi \leftarrow \alpha e^{-\beta t}$ \\
                        \hspace*{3.0em} Append current transition's $u_{epistemic}$ to $u_{history}$ \\
                        \hspace*{3.0em} Calculate: $moving\_avg_{unc} \leftarrow \frac{sum(unc_{history})}{len(unc_{history})}$ \\
                        \hspace*{3.0em} \textbf{if} $t \mathbin{\%} I_{\beta} == 0$ and $len(u_{history}) \geq N_{\text{window}}$ \textbf{then} \\
                            \hspace*{4.5em} $\beta \leftarrow \beta\ \times\ (\textbf{if}\ moving\_avg_{unc} > \xi\ \textbf{then}\ \beta_{dec}\ \textbf{else}\ \beta_{inc}\ \textbf{end if}$) \\
                        \hspace*{3.0em} \textbf{end if} \\
                        \hspace*{3.0em} \textbf{if} $t \mathbin{\%} I_{\alpha} == 0$ and $t > 0$ \textbf{then} \\
                            \hspace*{4.5em} $\alpha \leftarrow Q_{0.99}(u_{history})$ \\
                        \hspace*{3.0em} \textbf{end if} \\
                        \hspace*{3.0em} \textbf{if} $u_{epistemic} > \xi$ and $r > 0$ \textbf{then} \\
                            \hspace*{4.5em} Store the transition $\tau$ in the replay buffer $\mathcal{D}$, again \\
                        \hspace*{3.0em} \textbf{end if} \\
                \hspace*{1.5em} \textbf{end if} \\
            \textbf{end if} \\
    \end{flushleft}
    \vskip 3pt \hrule
\end{figure*}

\subsection{Loss Functions}
\label{sec:losses}

To train our agent model, we adopt the loss functions employed in the CEQR-DQN framework~\cite{stutts2024echoes-08b}. We adopt the CEQR-DQN framework as it uses efficient evidential deep learning to estimate uncertainty and the architecture is suitable for the framework proposed in this paper. 
We calculate the quantile Huber loss by combining quantile regression loss~\cite{koenker1978regression-0f3} with Huber loss~\cite{huber1964robust-ef0}. Conformalized joint prediction (CJP) can then be used to jointly train point predictions and uncertainty estimations~\cite{stutts2023lightweight-6f4}. The loss term $\mathcal{L}_{cal}$ is used to calibrate quantiles of the DQN target: 
\begin{equation}
    \mathcal{L}_{cal} = (1 - \lambda_{cal}) \times COV_{obj} + \lambda \times SHARP_{obj}
\end{equation}
where $\lambda_{cal}$ is the hyperparameter used to balance the coverage of $y$ with minimizing the length of the prediction interval, and: 
\begin{equation}
    SHARP_{obj} = \frac{1}{N} \sum_{i=1}^{N} \mathbb{I}\{q_i \leq 0.5\}[(1 - \hat{q}_i) - \hat{q}_i] + \mathbb{I}\{q_i > 0.5\}[\hat{q}_i - (1 - \hat{q}_i)]  
\end{equation}
\begin{equation}
    COV_{obj} = \frac{1}{N} \sum_{i=1}^{N}\mathbb{I}\{p^{cov}_{avg} < q_i\}[(y_i - \hat{q}_i)\mathbb{I}\{y_i > \hat{q}_i\}]  +\ \mathbb{I}\{p^{cov}_{avg} > q_i\} [(\hat{q}_i - y_i)\mathbb{I}\{y_i < \hat{q}_i\}]
\end{equation}
where $q$ is the target quantile and $\hat{q}_i$ is the predicted quantile.

EDL methods are considered to be heuristic and several research papers focus on mitigating this by using mathematically rigorous frameworks~\cite{gao2024comprehensive-b48,li2023survey-216,liu2026generalized-974,meinert2022unreasonable-a04}. Stutts et al.~\cite{stutts2024echoes-08b} calibrate both the target distribution and the mean of the evidential distribution jointly. For the target distribution, marginal coverage rate, $p_m$, is $0.5$. For the evidential distribution, the marginal coverage rate is $0.9$. A modified interval score function is then used for the evidential distribution: 
\begin{equation}
    \mathcal{L}_{interval} = \frac{1}{N} \sum_{i=1}^N (\hat{q_i}^{95^{th}} - \hat{q_i}^{5^{th}}) + \mathbb{I}\{y < \hat{q_i}^{5^{th}}\}\frac{2}{q}(\hat{q_i}^{5^{th}} - y) + \mathbb{I}\{y > \hat{q_i}^{95^{th}}\}\frac{2}{q}(y - \hat{q_i}^{95^{th}})
\end{equation}
where $\hat{q_i}$ is the predicted quantile. 
The evidential quantile regression loss is given by the following:
\begin{equation}
    \mathcal{L}_{evi} = \mathcal{L}_{NLL} + \lambda\mathcal{L}_{reg}
\end{equation}
where $\lambda$ is a tunable hyperparameter. $\mathcal{L}_{NLL}$ is the NLL loss function~\cite{amini2020deep-c4c} defined as follows:
\begin{equation}
    \mathcal{L}_{NLL} = \frac{1}{2}log\Bigg(\frac{\pi}{\nu}\Bigg) - \alpha log(\Omega) + \Bigg(\alpha + \frac{1}{2}\Bigg)log((y - \gamma)^2\nu + \Omega) + log\Bigg(\frac{\Gamma(\alpha)}{\Gamma(\alpha + \frac{1}{2})}\Bigg)
\end{equation}
and $\mathcal{L}_{Reg}$ is the regularization loss term:
\begin{equation}
    \mathcal{L}_{reg} = \phi[\mathbb{I}(y \geq \hat{y})q(y - \hat{y}) + \mathbb{I}(y < \hat{y})(1 - q)(\hat{y} - y)] . 
\end{equation}
The total evidential loss to be minimized is defined as: ${\mathcal{L} = \mathcal{L}_{evi} + \mathcal{L}_{cal} + \mathcal{L}_{interval}}$. For detailed derivations, we refer to Stutts et al. (2023)~\cite{stutts2023lightweight-6f4} and Stutts et al. (2024)~\cite{stutts2024echoes-08b}. 

\begin{figure}[!t]
  \centering
  \begin{minipage}[t]{0.3\textwidth}
    \vspace{0pt}
    \includegraphics[width=\linewidth]{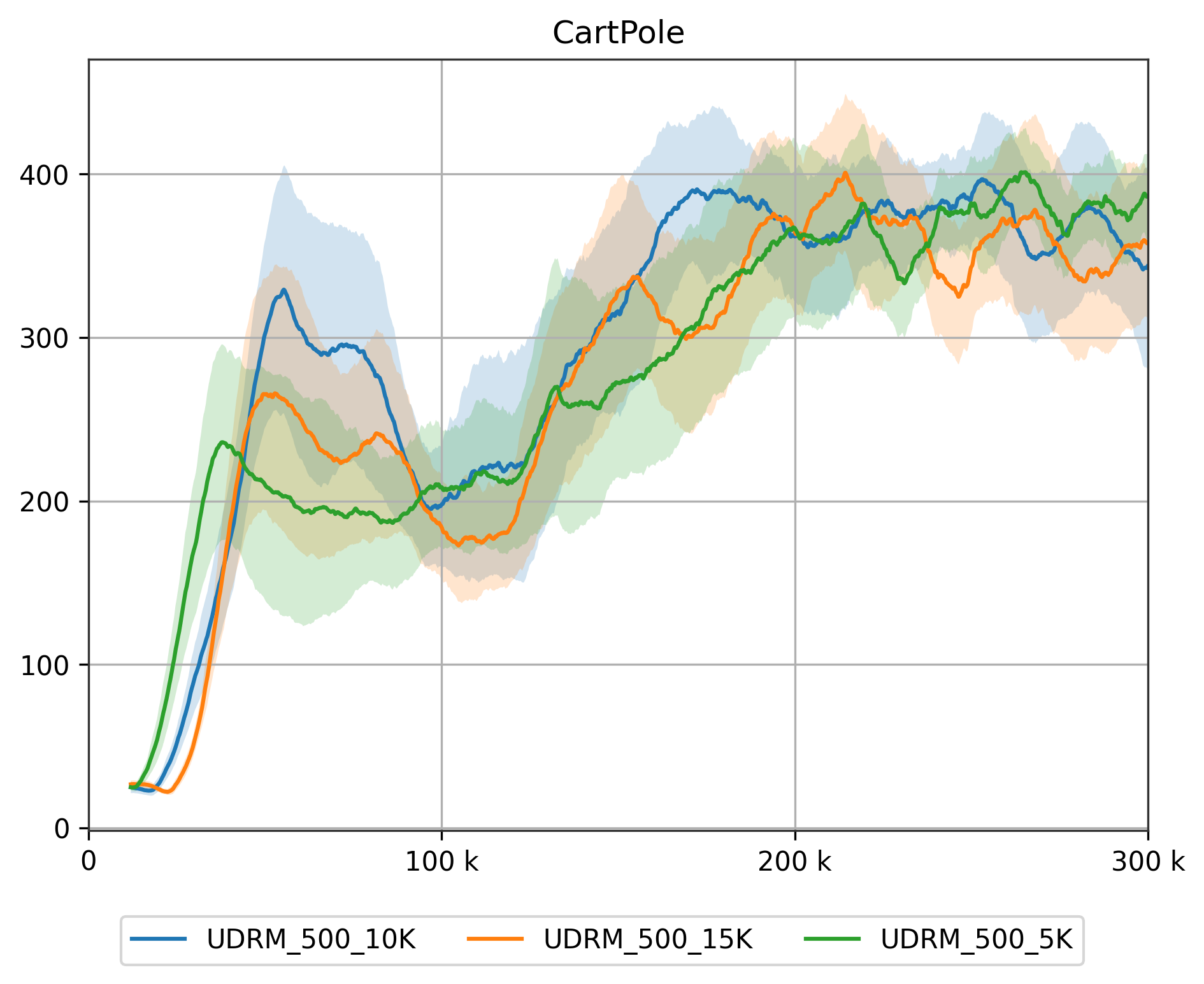}
    \captionof{figure}{Performance of CartPole for different pre-training steps. Lower number of steps appear to have better performance. \label{fig:cp_train}}
  \end{minipage}%
  \hspace{0.5cm}%
  \begin{minipage}[t]{0.52\textwidth}
    \vspace{0pt}
    \centering
    \begin{tabular}{>{\raggedright\arraybackslash}p{0.43\linewidth} >{\centering\arraybackslash}p{0.59\linewidth}}
        \toprule
        \textbf{Hyperparameter} & \textbf{UDRM} \\
        \midrule
        Batch Size & $32$ \\
        N Quantiles & $50$ \\
        Replay Buffer Capacity & $100000$ \\
        Training Start & $5000$ \\
        Target Update Frequency & $1000$ \\ 
        Discount Factor & $0.99$ \\
        Training steps & $2500000$ \\
        Learning Rate & $10^{-4}$ \\
        Adam $\epsilon$ & $10^{-4}$ \\
        Update Frequency & $1$ \\
        Kappa & $1$ \\
        $\lambda_{al}$ & $0$ \\
        $\lambda_{ep}$ & $\{0.001, 0.01, 0.0002, 0.005, 0.0005\}$ \\
        \bottomrule
    \end{tabular}
    \captionof{table}{UDRM - Training hyperparameters for MinAtar \label{tab:hyperparameters}}
  \end{minipage}
\end{figure}

\section{Experimental Results}
\label{sec:experiments}


In this section, we discuss the performance of the proposed model, UDRM, against other baseline models, on the MinAtar testbed~\cite{young2019minatar-eaa}. This is a miniaturized version of the Atari games that is used for evaluating several RL baseline models. We use the MinAtar testbed for evaluation since the baseline methods, i.e., UA-DQN~\cite{clements2019estimating-07f} and CEQR-DQN~\cite{stutts2024echoes-08b}, evaluate against this testbed in their respective prior efforts. 

We compare the performance of our proposed framework, UDRM, with UADQN~\cite{clements2019estimating-07f} and CEQR-DQN~\cite{stutts2024echoes-08b}, particularly since we have retained the RL agent's network structure from these models. UDRM is evaluated with two $\alpha$ recalibration intervals, i.e., $\alpha=50$ and $500$. For each baseline model and environment benchmark, we ran experiments for $10$ uniquely seeded trials across $2.5$ million time steps on the MinAtar games (Asterix, Breakout, Freeway, Seaquest, Space Invaders). All experiments were run on one A$100$ GPU. Classic control and toy text environments were trained for $10$ seeds in about $12$ hours. MinAtar environments required $12$ to $15$ hours to train one seed. Table~\ref{tab:avg_cum_reward} reports mean and standard deviation of average cumulative reward for MinAtar games and Table~\ref{tab:classic_toy_reward} reports the same for classic control and toy text environments.

\begin{figure*}[!ht]
    \centering
    \includegraphics[width=\textwidth]{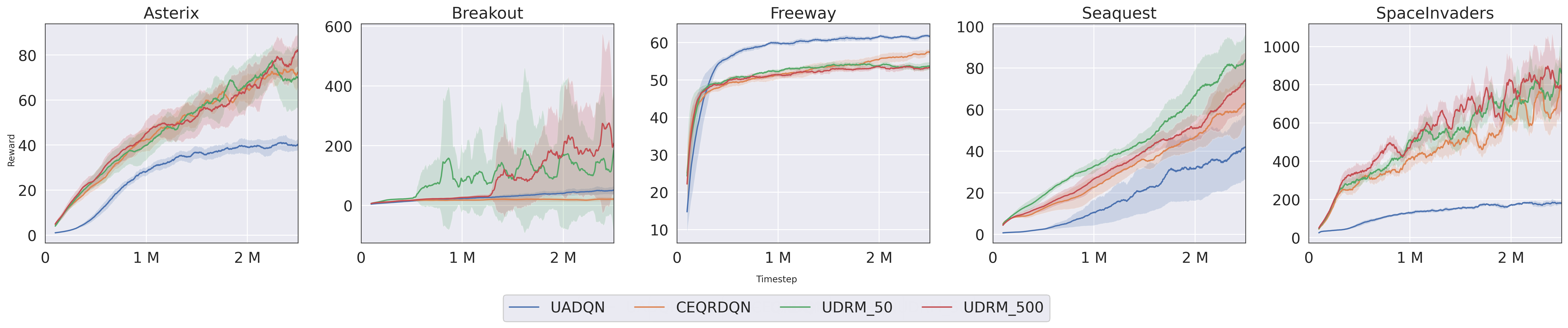}
    \caption{Performance of UADQN, CEQR-DQN, and UDRM (for two $\alpha$-recalib intervals $50$, $500$ and $\beta$-recalib interval $5000$) on MinAtar. UDRM achieves higher rewards for Asterix, Breakout, Seaquest, and SpaceInvaders. For Freeway, UDRM is competitive with other baseline models.}
    \label{fig:atari_results}
\end{figure*}

Our proposed UDRM framework achieves a higher maximum score compared to UADQN and CEQR-DQN, as shown in Figure~\ref{fig:atari_results}, in the Asterix, Breakout, Seaquest, and SpaceInvaders environments. In most games, UDRM achieves higher scores at each time step compared to other models. For Asterix and Breakout, UDRM, with an alpha recalibration interval of $500$ time steps, performs better than other models while for Seaquest and SpaceInvaders, UDRM with an alpha recalibration interval of $50$ time steps performs better. For Freeway, UDRM is competitive against CEQR-DQN.\footnote{Freeway is a sparse-reward environment in which simpler models perform better.} 
Note that the proposed model, UDRM, performs better than other baseline models for four out of five MinAtar games. Figure~\ref{fig:cp_train} shows an ablation study performed for a limited set of values of pre-training step counts, on the CartPole environment. Observe that a smaller pre-training phase appears to enable the agent to achieve greater rewards.

\begin{figure*}[!ht]
    \centering
    \includegraphics[width=0.9\textwidth]{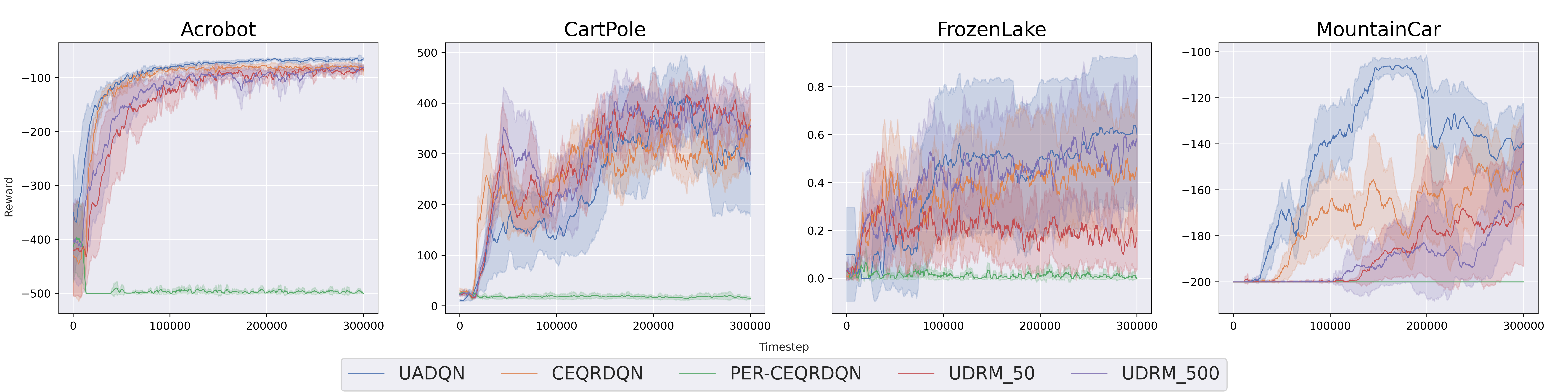}
    \caption{Performance of UADQN, CEQR-DQN, CEQR-DQN (with PER), and UDRM (for two $\alpha$-recalib intervals $50$, $500$ and $\beta$-recalib interval $5000$) on Classic Control and Toy Text environments. UDRM is competitive against other models in these smaller environments.}
    \label{fig:classic_toy}
\end{figure*}

Figure~\ref{fig:classic_toy} shows the performance of UDRM in classic control and toy text environments. UDRM learns as well as CEQR-DQN on the classic control and toy text environments. UADQN performs better than the other models in three out of four of these smaller environments. We also compare our proposed method's performance against prioritized experience replay~\cite{schaul2015prioritized-85b}, modified to use pairwise Quantile Huber loss~\cite{dabney2018distributional-0e1, millard2026federated-22b, niu2021di-engine-56f} as the prioritization metric. Experiments show that this PER variant performs the worst out of all models, particularly in the classic control and toy text environments. This can be attributed to the fact that PER prioritizes transitions based on the TD error and initial training steps populate the replay memory buffer with transitions having high TD errors. Training on such transition samples proves to be inefficient as these are, in effect, unlearnable transitions. The minimum scores for UDRM are comparable to those of CEQR-DQN.\\
\indent Future work should consider improving UDRM to reduce memory requirements and designing yet more efficient methods to calculate and update the uncertainty threshold. The training hyper-parameters may be environment-dependent; thus, further tuning of these values may unlock additional performance gains. Table~\ref{tab:hyperparameters} shows the hyper-parameters we used for the experiments of this study.

\section{Conclusions}
\label{sec:conclusions}

In this work, we propose, for training reinforcement learning (RL) control agents, uncertainty-driven replay memory (UDRM), an experience replay memory buffer that is populated based on estimated uncertainty values provided by the model/controller during training. The UDRM buffer enables the agent to take better actions (i.e., make better decisions) by leveraging uncertainty scores to bias the storage of transitions, that the agent is uncertain about, into its experiential memory. This process skews the memory's maintained distribution of transitions towards uncertain actions which ultimately improve agent exploration and generalization ability. Empirical results demonstrate that our UDRM framework achieves better or competitive performance asmcompared to other uncertainty-aware baseline models, including an augmented prioritized experience replay, across several tasks.



\bibliographystyle{acm}
\bibliography{refs}

\end{document}